\documentclass[letterpaper, 10 pt, conference]{ieeeconf}  

\IEEEoverridecommandlockouts                              

\usepackage{lipsum}
\usepackage{graphicx}
\usepackage{graphics}
\usepackage{amsfonts}
\usepackage{siunitx}
\usepackage{float}
\usepackage{hyperref}
\usepackage{epstopdf}
\usepackage{enumerate}
\usepackage{cite}
\usepackage{xcolor}
\usepackage{amsmath}
\usepackage{bm}
\hypersetup{hidelinks} 
\graphicspath{{figures/}}
\usepackage{balance}

\newcommand{\bmd}[1]{\bm{\dot{#1}}}

\def\equationautorefname~#1\null{(#1)\null}
\def\figureautorefname~#1\null{Fig. #1\null}

\title{\LARGE \bf
Fine Manipulation and Dynamic Interaction in Haptic Teleoperation
}

\author{Carlo Tiseo$^\dagger$, Quentin Rouxel$^\dagger$, Zhibin Li, and Michael Mistry 
\thanks{$^\dagger$ The authors contributed equally to this paper.}
\thanks{This work has been supported by EPSRC UK RAI Hub ORCA (EP/R026173/1), the Future AI and Robotics for Space (EP/R026092/1), National Centre for Nuclear Robotics (NCNR EPR02572X/1) and THING project in the EU Horizon 2020 (ICT-2017-1).}
\thanks{Carlo Tiseo is with the School of Engineering and Informatics, University of Sussex (Chichester 1 Room 002, Brighton BN1 9QJ, UK). All the authors are with the ECR, IPAB, School of Informatics, University of Edinburgh (10 Crichton St, Edinburgh, EH8 9AB, UK).{\tt\small c.tiseo@sussex.ac.uk, quentin.rouxel@ed.ac.uk}}%
}

\begin{document}
\maketitle
\thispagestyle{empty}
\pagestyle{empty}

\begin{abstract}
The teleoperation of robots enables remote intervention in distant and dangerous tasks without putting the operator in harm's way. However, remote operation faces fundamental challenges due to limits in communication delays. The proposed work improves the performances of teleoperation architecture based on Fractal Impedance Controller (FIC) by integrating into the haptic teleoperation pipeline a postural optimisation that also accounts for the replica robots' physical limitations. This update improves dynamic interactions by trading off tracking accuracy to maintain the system within its power limits. Thus, allowing fine manipulation without renouncing the robustness of the FIC controller. Additionally, the proposed method allows an online trade-off between tracking the autonomous trajectory and executing the teleoperated command, allowing their safe superimposition. The validated experimental results show that the proposed method is robust to increased communication delays. Moreover, we demonstrated that the remote teleoperated robot remains stable and safe to interact with, even when the communication with the master side is abruptly interrupted.
\end{abstract}

\maketitle

\section{Introduction}
Teleoperation finds application in medical, space and industrial robotics. It allows remote intervention, which can be helpful to reduce the time required to complete a task, keep operators away from a dangerous situation, and enable the accessibility to a service that would be otherwise prohibitive \cite{schmaus2019knowledge,toet2020toward,penco2018robust}. For example, the teleoperated system can provide assistance and medical treatment remotely, allowing patients to access specialised medical operators that would be otherwise impossible to intervene. Addressing all these applications requires a modular adaptive control architecture, which is capable of fine manipulation, dynamic interaction, and robust and safe to operate even when there is a loss of connection between the master and the replica device. 

Over the last few decades, there has been a steep increase in performances in teleoperated systems. Commercial devices targeting medical, military and civil applications are now available; however, their manipulation capabilities are often limited or constrained to specific types of interactions \cite{chen2019rbf,lipton2017baxter,hulin2021model,elobaid2019telexistence}. Furthermore, these systems usually require extensive tuning to be re-purposed for new applications, limiting the applicability of these systems in complex tasks that might need, for example combining fine manipulation and dexterous dynamic interaction in an unstructured environment \cite{schmaus2019knowledge,feizi2021robotics,beckerle2019robotic}. These problems are also amplified when dealing with haptic teleoperation, introducing additional complexity into the control architecture that might become more fragile to delays between the two systems \cite{chen2019adaptive,su2019neural,su2020deep}.

A few recently proposed architectures have shown the capabilities of passive controllers in performing haptic teleoperation and manipulation in an unstructured environment \cite{babarahmati2020,Babarahmati2021TeleCoop}. Passivity can be either an inherit property of the controller, or a controller can be made passive using an energy-tank \cite{babarahmati2019,ferraguti2015,minelli2019energy}. Tank-based methods are limited by the presence of residual energy in the tank; therefore, the controller performance are compromised in the occurrence that the tank runs out of energy. Furthermore, the approach proposed in \cite{minelli2019energy} for multi-arm teleoperation requires all the master and replica systems involved to share the same energy reservoir, which increases the computational complexity of the architecture and couples the stability of the robots' controllers. 

On the other hand, the methods proposed in \cite{babarahmati2020} and \cite{Babarahmati2021TeleCoop} are based on an intrinsically passive controller, exploiting the stability proprieties of the Fractal Impedance Controller (FIC) theoretically proved in \cite{babarahmati2019,Tiseo2020Bio}. Their performances show that these FIC-based methods can exploit highly non-linear stiffness profiles to generate regions of compliant behaviour without compromising the robot tracking accuracy even in the presence of communication delays. The FIC also allows to decouple the stability of the different robots and can be tuned online \cite{babarahmati2020}, \cite{Babarahmati2021TeleCoop}. Thus, the system can adapt online to different tasks' requirements, perform autonomous haptic exploration and mimic human behaviour during dynamic interaction \cite{Tiseo2021MIGame,Tiseo2021HapFic,Tiseo2021H-FIC}. Some of the task that have been tested are driving a pile into sand, pushing a ball across a highly irregular terrain, drilling hole patterns on different materials, and cooperating with a human during unstructured tasks \cite{babarahmati2020,Babarahmati2021TeleCoop}. More recently, \cite{Tiseo2021MIGame} proposed an FIC-based architecture capable of mimicking human movements in dexterous tasks such as drawing and writing under uncertain interaction. 

\begin{figure*}[thb]
      \centering
      \includegraphics[width=\textwidth, trim= 2.75cm 21.75cm 2.5cm 0cm,clip]{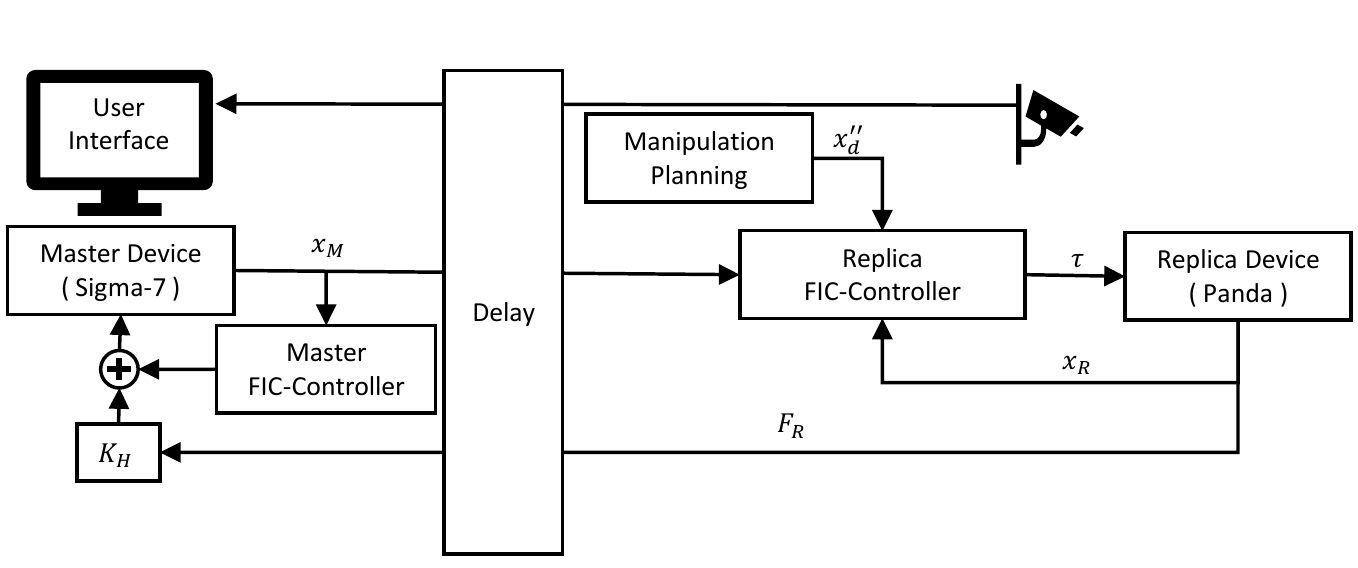}
      \caption{User receives visual feedback from two fixed cameras pointing at the workspace and haptic feedback from the master device. Haptic force feedback produced by the Master combines the force at the Replica end effector scaled by a factor $K_\text{H}$, and the master FIC controller providing force feedback about the workspace boundaries determined by the Replica FIC-Controller. On the Replica side, the desired position of the manipulation planning ($x_\text{d}$) is combined with the user input ($x_\text{M}^{\prime\prime}$) to generate the torque command ($\tau$) sent to the Replica.}
      \label{fig:Architecture}
\end{figure*}

In summary, FIC architectures have shown that this controller renders the robot capable of dexterous robust interaction. Earlier architectures have been tested with feedback delays up to \SI{1}{\second} and controller bandwidth as low as \SI{10}{\hertz} both in manipulation and teleoperation \cite{babarahmati2019,babarahmati2020}. Previous FIC architectures have also shown that switching between teleoperation and manipulation is possible without any tuning. However, they required restarting the system to change the controller modality; there was no postural optimisation. The controller was only able to follow discrete sequences via points and not trajectories, and it could trigger the protection mechanisms (i.e., velocity limits and torque rate) of the robot when there was a sudden acceleration due to the lack of constraints on the controller power limits. In this manuscript, we refine and integrate our earlier methods into a single architecture (\autoref{fig:Architecture}). This new controller addresses the following points:
\begin{enumerate}
    \item The proposed architecture increases the system remains operational during delayed haptic teleoperation by autonomously adjusting the robot behaviour to maintain the control commands within the system physical constraints and avoid triggering the robot protection mechanisms (i.e., velocity limits and torque rate) by trading off its tracking performance. The system has been tested for communication delays of about $\SI{200}{\milli\second}$, implying a feedback latency of $\SI{400}{\milli\second}$.
    \item Validation that the updated replica controller on the robot allows the superimposition of teleoperated behaviour and autonomous motion by adding together these two motion components without affecting the system stability.
\end{enumerate}

The control architecture is presented in \autoref{sec:ControlArchitecture}. The experimental validation is described in \autoref{sec:ExperimentalValidation}, and they are discussed in \autoref{sec:Discussion}.

\section{Control Architecture}\label{sec:ControlArchitecture}
The architecture in \autoref{fig:Architecture} has a hierarchical structure of semi-autonomous controllers, which allow controlling the replica device. The master device is used to measure the operator's movements while providing online adjustable haptic feedback. The replica controller is composed of two parallel FIC-based control that take as input the output of the manipulation planner superimposed to the position offset recorded on the master device. Additionally, the teleoperation controller also has a velocity-based command, enabling to move the equilibrium position of the replica controller to re-centre the system workspace manually. The operator can switch between the offset and the velocity based teleoperation using a keyboard command.

\subsection{Monodimensional Fractal Impedance Controller}

The task-space control uses the FIC formulation using the force profile introduced in \cite{tiseo2020Planner}, and later redefined in \cite{Tiseo2021H-FIC}, which is defined as follows for each dimension:
\begin{equation}
    \label{eq:FICForce}
    \begin{array}{l}
        F_\text{c}(\tilde{x}) = \\\left\{\begin{array}{ll}
        K_0\tilde{x}, & \|\tilde{x}\|\le \xi \tilde{x}_\text{b} \\
        \cfrac{\Delta F}{2}\left(\tanh\left(\cfrac{\tilde{x}-\tilde{x}_\text{b}}{S\tilde{x}_b}+\pi\right)+1\right)+F_0, & \text{else}.
    \end{array}\right.    
    \end{array}
\end{equation}
$K_0$ is the constant stiffness, $\tilde{x}=x_\text{d}-x$ is the end-effector position error, and $\tilde{x}_\text{b}$ is the tracking error when the force saturation occurs. Being $F_\text{max}$ the saturation of the force of the controller, $\Delta{F}=F_\text{max}-F_0$, $F_0=\xi K_0\tilde{x}_\text{b}$, $S=\left(1- \xi
\right)\left(\tilde{x}_\text{b}/(2\pi)\right)$ controls the saturation speed, and $\xi \in [0,~1]$ control the starting of the saturation behaviour while approaching $x_b$. The $F_\text{c}$ is then substituted in the attractor formulation.
\begin{equation}
    \label{eqFICattractor}
    FIC(\tilde{x})=\left\{\begin{array}{cc}
        F_\text{c}(\tilde{x}),  & \text{D} \\
        \cfrac{2F_\text{c}(\tilde{x}_\text{max})}{\tilde{x}_\text{max}}\left(\tilde{x}-\cfrac{\tilde{x}_\text{max}}{2}\right) & \text{C},
    \end{array}\right.
\end{equation}
where (D) indicates the divergence phase, and (C) indicates the convergence phase, as defined in \cite{babarahmati2019}. $\tilde{x}_\text{max}$ is the maximum state error recorded at the beginning of the last convergence phase. The convergence phase is identified using the following conditions on the system state: $\|\tilde{x}(t)\|<\|\tilde{x}(t-1)\|$ and $\mathit{sign}(\tilde{x}(t))\ne \mathit{sign}(\tilde{x}(t-1))$.

The FIC is an intrinsically stable controller being the algorithmic implementation of a passive non-linear oscillator with autonomous harmonic trajectories converging to the desired state \cite{babarahmati2019,Tiseo2021HapFic,Tiseo2021H-FIC}. Its autonomous trajectories also upper bound the energy and the power that the controller releases, which can be experimentally calibrated to the system to guarantee stability \cite{babarahmati2019,Tiseo2020Bio}. Another benefit of this controller is that its observer coincides with its non-linear spring. This implies that being conservative, it is path independent and robust to low controller bandwidth, and feedback delays \cite{babarahmati2019,babarahmati2020}. Additionally, passive and conservative systems can be superimposed without affecting their stability \cite{babarahmati2020}, allowing to decouple the stability of the master and the replica robots in the proposed architecture.

The sole requirement to guarantee the stability of the FIC is that the used force profile is continuous with bounded derivatives (i.e., Lipschitz function) as shown in the Lyapunov stability analyses presented in \cite{babarahmati2019,Tiseo2020Bio}. The proposed force profile in Eq. \autoref{eq:FICForce} is a smooth continuous function, which satisfies the stability condition of the FIC controllers.

\subsection{Master Controller}
The master controller also refines the earlier formulations presented in \cite{babarahmati2019} and \cite{babarahmati2020}. The first difference is switching to the FIC algorithm introduced in \cite{Tiseo2020Bio,Tiseo2021HapFic}, which removes the jump in force when switching from divergence to convergence. The second is giving the operator a direct control on the haptic feedback scaling factor ($K_\text{H}$). This function was implemented using the grasp degrees of freedom of the Sigma-7 as an analogue input. The online tuning of the haptic feedback is required to modulate the system responsiveness to the external interactions. Due to unbalanced inertia between the master and replica device, the interaction efforts measured on the replica will generate a higher twist on the master than the latter. This can be addressed by introducing a fixed scaling factor on the recorded forces. However, this approach also renders small interaction forces, which might be informative for some tasks, undetectable to the user interacting with the master. Therefore, we opted for online tuning of the scaling (using the grasp-dof of the Sigma-7) to enable the user to have an online control of the system responsiveness. 

The operator commands the task space motion with respect to a reference equilibrium position $\bm{x}_\text{Md}$ defined as the local origin $\bm{x}_\text{Md} = \bm{0}$.
\begin{equation}
    \bm{f} = \bm{FIC}(-\bm{x}_\text{M}) + K_\text{H}\bm{F}_\text{R},
\end{equation}
where $\bm{f}$ is the end-effector control effort of the master device, $\bm{x}_\text{M}$ is the end-effector pose of the master device. $K_\text{H}\in \left[0,~1\right]$ is the haptic gain controlled online by the user with the grasp-DoF of the Sigma-7 device.  $\bm{F}_\text{R}$ is the force measured at the end-effector of the replica. 

The operator can choose between two control modalities for teleoperating the replica robot, which alter the interpretation of the signal $\bm{x_\text{M}}$ by the Replica FIC-Controller (\autoref{fig:Architecture}). The Offset-Based Teleoperation ($\bm{x}^{\prime}_\text{d} = \bm{x}_\text{M}$) gives better dexterity, but it is limited to the workspace determined by the Master range of motion. The second modality is a Velocity-Based Teleoperation ($\bm{x}^{\prime}_\text{d}(t) =\bm{x}^{\prime}_\text{d}(t-\Delta t)+\bm{x}_\text{M}\Delta t$), which allows a broader range of motion, sacrificing some dexterity.

\subsection{Trajectory Planning for Manipulation}
The trajectory planning exploits the FIC architecture presented in \cite{Tiseo2021MIGame}, which improves our earlier results presented in \cite{tiseo2020Planner,tiseo2021,Tiseo2021H-FIC}. The main difference of the updated formulation is the ability to track trajectories and not only discrete sequences of via points. 

The formulation used for the planner is:
\begin{equation}
    \label{eq:HarmonicTrajectoryPlanner}
    \begin{array}{l}
         \bm{x}_\text{d}^{\prime\prime}=\displaystyle{\int_0^t \dot{\bm{x}}_\text{d}^{\prime\prime}~dt} \\\\
        \dot{\bm{x}}_{d}^{\prime\prime}=\displaystyle{\int_0^t \ddot{\bm{x}}_\text{d}^{\prime\prime}~dt~~~ \in [-\bm{v}_\text{d},\bm{v}_\text{d}]}\\\\
        \ddot{\bm{x}}_\text{d}^{\prime\prime}=\left\{\begin{array}{l}
                 \mathrm{sign}\left(\tilde{\bm{x}}_\text{t}\right)\min\left(\omega_n^2\left\|\tilde{\bm{x}}_\text{t}\right\|,\|\bm{a}_\text{max}\|\right)-\mu\dot{\bm{x}}_\text{d}^{\prime\prime}, ~\text{D}\\
                 \cfrac{2A_\text{max}}{\tilde{\bm{x}}_{\text{T}0}}\left( \bm{x}_\text{d}^{\prime\prime}\left(t-1\right)-\cfrac{\tilde{\bm{x}}_{\text{T}0}}{2}\right)-\mu\dot{\bm{x}}_\text{d}^{\prime\prime},  ~~~\text{C},
              \end{array}\right.
    \end{array}
\end{equation}
where $\tilde{\bm{x}}_\text{t}(t)=\bm{x}_\text{t}(t)-\bm{x}_\text{d}^{\prime\prime}(t-1)$, $\bm{x}_\text{t}(t)$ is the next point in the desired trajectory, $\omega_n$ is the desired natural frequency of the planner in \si{\radian\per\second}, and $\mu=0.01\omega_n$ is the viscosity. In regards of the equation for the (C) phase, $A_\text{max}$ is the acceleration associated with the maximum displacement ($\tilde{\bm{x}}_{\text{T}0}$) reached in the previous (D) phase. $\bm{a}_\text{max}$  describes the maximum acceleration limit of the movement and it is derived from the desired tangential velocity $v_\text{d}$ as follows:
\begin{equation}
    \label{eq:HarmonicTrajectoryPlannerParameter}
    \begin{array}{l l}
             v_\text{max}=v_\text{p}\min\left(v_\text{d},~\omega_n ||\bm{d}|| \right), &
            \bm{a}_\text{max}=2\bm{d}\left(\cfrac{v_\text{max}}{||\bm{d}||}\right)^2,
    \end{array}
\end{equation}
where $v_\text{p}=1.595$, $\bm{d}$ is the distance vector from $\bm{x}_\text{t}$ when the new command is issued.

\subsection{Replica Controller}
The torque control signal for the Replica robot is:
\begin{equation}
    \bm{\tau} =\bm{C}(\bm{q}, \bm{q})+\bm{G}(\bm{q})+\bm{J}(\bm{q})^{T}\left(FIC(\tilde{\bm{x}}_\text{R})\right),
\end{equation}

where $\bm{C}(\bm{q}, \bmd{q})$ and $\bm{G}(\bm{q})$ are the vectors of centrifugal, Coriolis and gravitational forces in joint space,
$\bm{J}(\bm{q})$ is the kinematic Jacobian of the end-effector, and $\tilde{\bm{x}}_\text{R}=\bm{x}_\text{d}-\bm{x}_\text{R}$ is the tracking error of the replica robot.
$\bm{x}_\text{d}=\bm{x}_d^{\prime}+\bm{x}_d^{\prime\prime}$ and $\bm{x}_R$ are the desired and measured respectively position the end-effector.

\section{Experimental Validation}\label{sec:ExperimentalValidation}
\begin{figure}[thb]
      \centering
      \includegraphics[width=\columnwidth, trim= 6cm 23.5cm 6cm .5cm,clip]{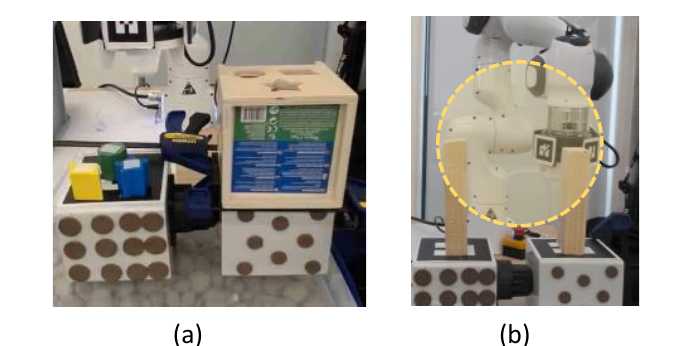}
      \caption{Experimental Tasks: (a) Shape sorting cube experiments involving a parallelepiped, an octagonal prism, and a star-shaped dodecagonal prism. (b) Manipulator tracks the circular trajectories in yellow, and the user teleoperates the manipulator to avoid or induce interaction with the two wood obstacles.}
      \label{fig:Experiments}
\end{figure}
The proposed method has been tested in the two experiments shown in \autoref{fig:Experiments} to evaluate the robustness of interaction and accuracy, which have been designed to validate the controller in ideal conditions (i.e., expert operator). The first experiment involved teleoperating the robot via the haptic interface to solve a shape sorting cube. For the second experiment, the manipulator is autonomously tracking a circular trajectory on a collision course with two obstacles, and the operator task is to avoid collisions between the robot and the two objects. We conducted these experiments with (one-way) communication delays of about \SI{200}{\milli\second} between the master and the replica robot, which we consider is enough to prevent the use of a haptic teleoperation without the introduction of shared autonomy algorithms in replica robots. Furthermore, the stability of the architecture was tested in the event of a loss of communication between the master and the slave robot. The FIC parameters used for the replica are $\tilde{x}_\text{b}=\SI{.05}{\meter}$, $K_0=\SI{200}{\newton\meter^{-1}}$, which implies that an error up to $\SI{.05}{\meter}$ is an acceptable tracking accuracy. It is worth noting that the controller can be make stiffer if needed as shown in \cite{Tiseo2021MIGame}.

\subsection{The Shape Sorting Cube}
The shape sorting cube is a child game training hand-eye coordination, fine motor and problem-solving skills. Consequently, this game is well suited to test the capability of the teleoperation setup of executing fine motor skills, which is also challenged by receiving limited visual feedback from only two cameras with a fixed point of view. Additionally, the experiment is designed to evaluate the architecture robustness to dynamic and unstructured interactions and its capability to pass from a force-based task to a fine motor skill without re-tuning the controller. This test is achieved by replacing the gripper with a Velcro strap attached to a static end-effector plate. These fastening tapes are constituted by two different tapes locking on each other. One of the tapes has semi-rigid hooks; the other tape has a soft pile of synthetic fabric. The degree of interlocking between the two sides determines the bonding strength. Thus, generating an inconsistent bound that is perfect for testing our controller's robustness and dynamic properties. In summary, the assigned task is: \textit{i}) Pick up one of the objects. \textit{ii}) Align and insert in the proper shape in the box. \textit{iii}) Lock the object against the hole geometry and pull. \textit{iv}) Break the Velcro strap bond and complete the task.

\begin{figure*}[thb]
      \centering
      \includegraphics[width=.9\textwidth, trim= 2cm 8cm 2cm 7cm,clip]{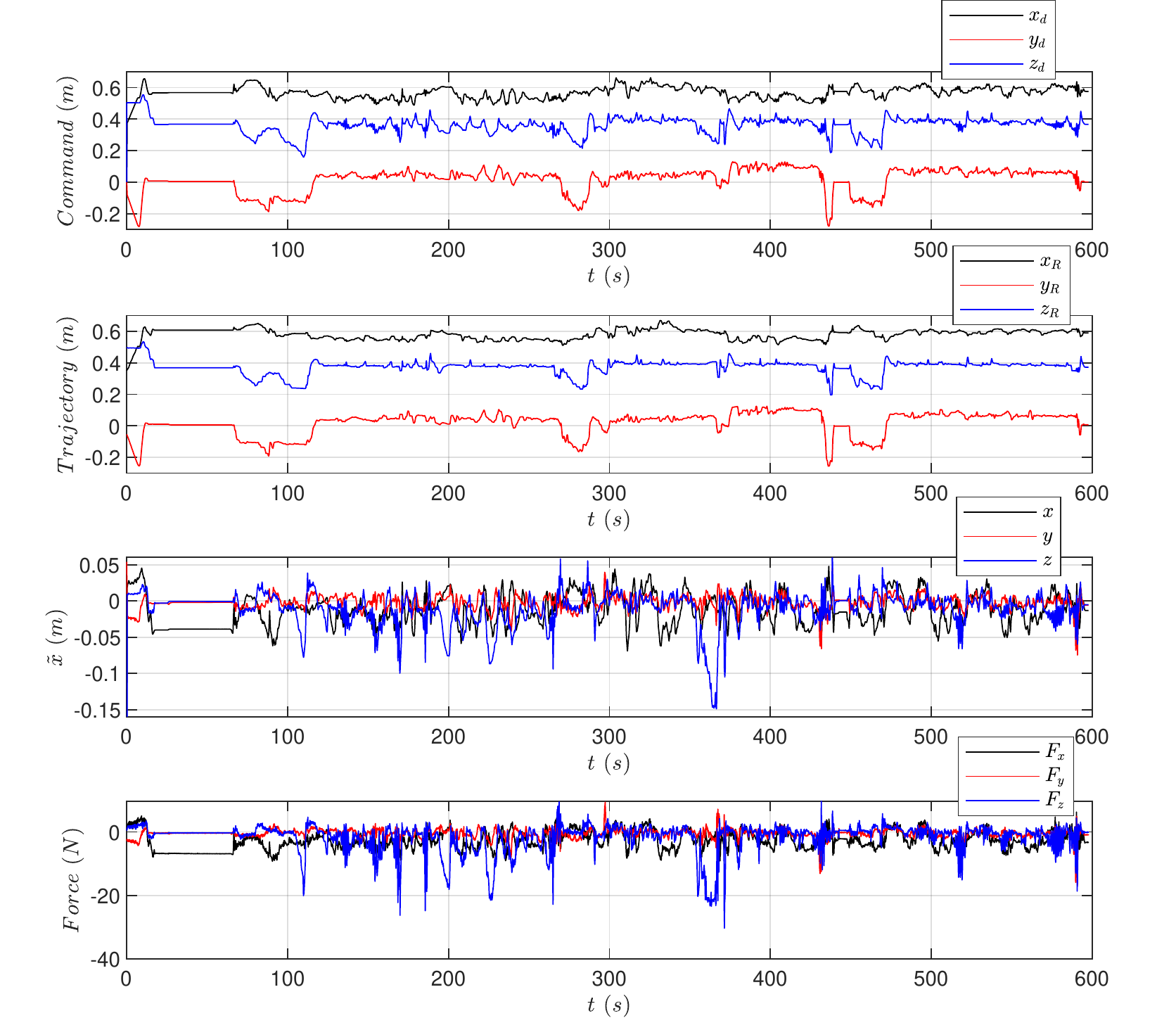}
      \caption{The data indicate that the tracing accuracy is within the selected level ($\tilde{x}_b=\SI{0.05}{\meter}$) whenever the system is not interacting with the environment. There are multiple peaks of the impacts that exceed the maximum value that can be generated by the controller, highlighting the dynamically challenging environment produced by these tasks. However, the controller can absorb the energy and maintain a safe and robust interaction.}
      \label{fig:TrajectoryExp1}
\end{figure*}

\begin{figure*}[thb]
      \centering
      \includegraphics[width=.9\textwidth, trim= 3cm 9.75cm 3cm 12.25cm,clip]{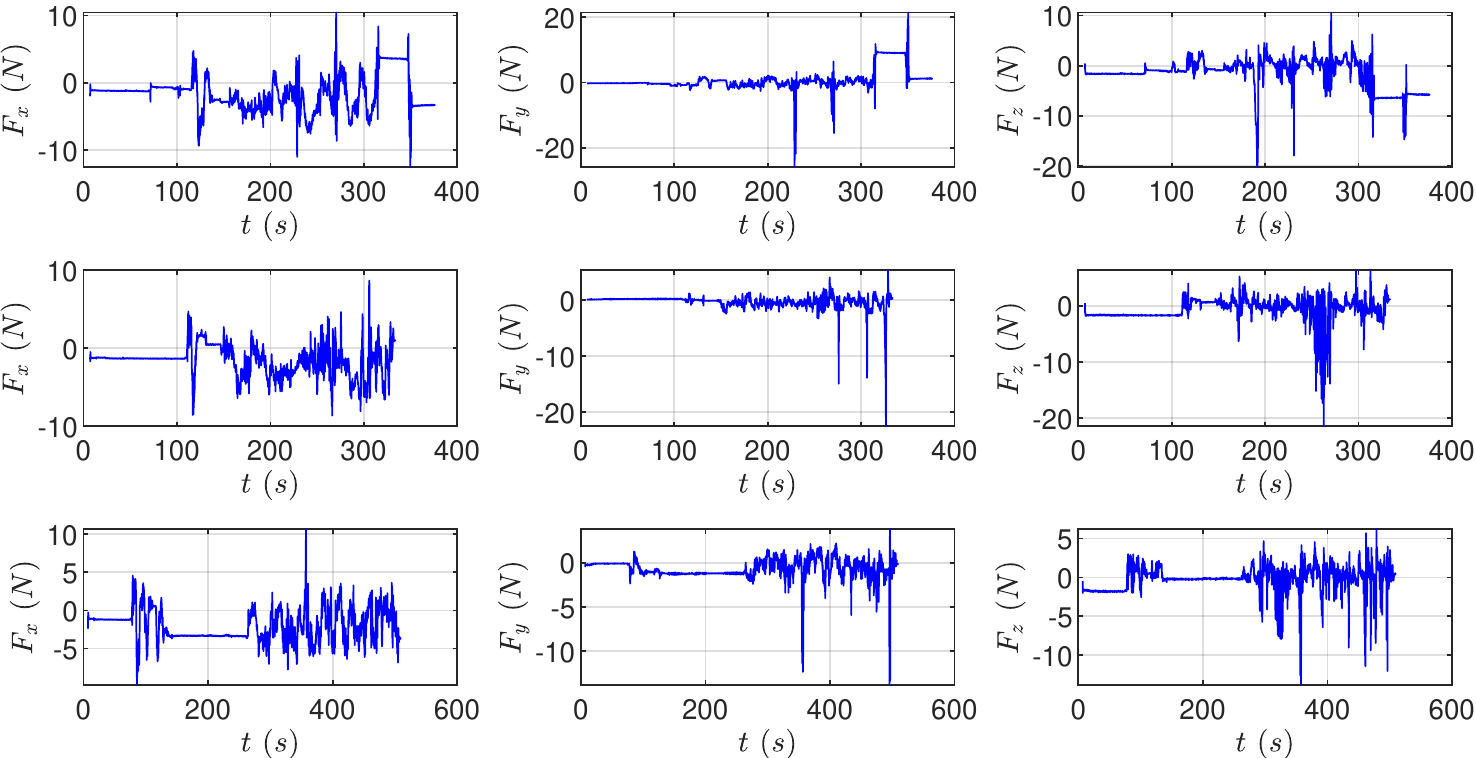}
      \caption{Interaction forces recorded during the other three trials of the shape sorting cube. The second row from the top is the trial testing for the robustness to loss of communication between Master and Replica. After the disconnection occurred at about \SI{280}{\second}, the replica continues to safely interact with the environment as in the other 2 experiments. The force signals show how the controller is intrinsically robust to multiple impulsive interactions with the environment.}
      \label{fig:InteractionForces}
\end{figure*}

The dynamic characteristics of the proposed method are highlighted in the experiments' video \cite{video}. \autoref{fig:TrajectoryExp1} shows the trajectory during the first shape sorting experiment. The tracking error is within the selected accuracy when the environmental interaction does not impede it. The force data also highlight how the controller can safely absorb interaction force that exceeds the maximum value that can be generated by the controller, highlighting the dynamically challenging environment generated by the selected experiment. The interaction forces recorded during the other 3 trials are shown in \autoref{fig:InteractionForces}, including an experiment that includes with a loss of communication between the master and the replica. Such events generate instability in all the teleoperation architectures that require observability of both master and replica states to guarantee the stability of interaction \cite{minelli2019energy}. The interaction forces recorded reach magnitudes consistently exceeding \SI{10}{\newton} with a maximum value of \SI{30.4}{\newton}. 

\begin{figure*}[thb]
      \centering
      \includegraphics[width=.9\textwidth, trim= 2cm 22cm 2cm 1cm,clip]{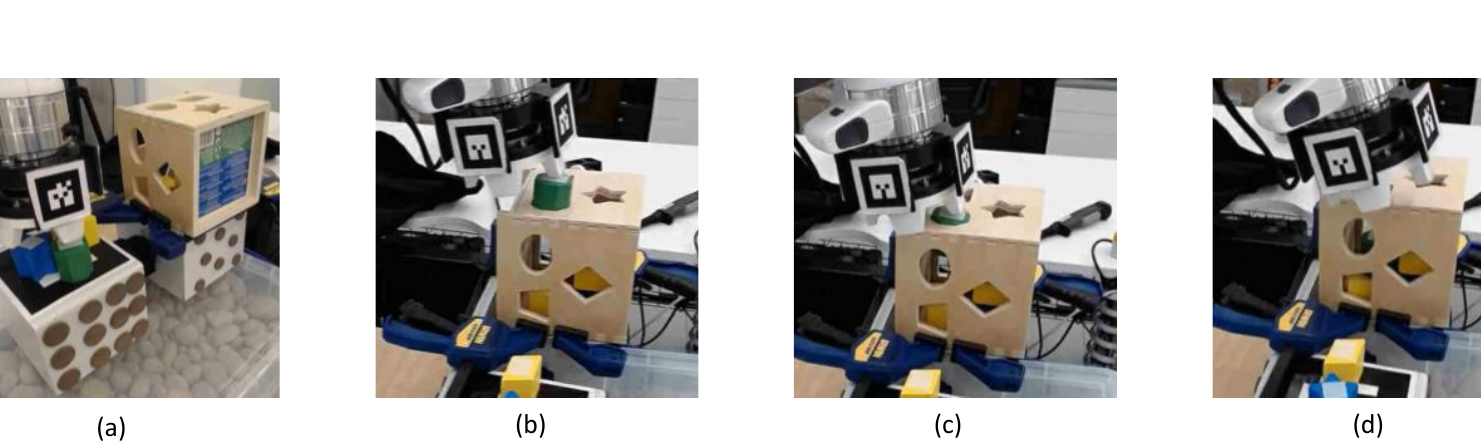}
      \caption{Task's stages for the first experiment: (a) Object is picked up making a bound using the Velcro strap. (b) Object is aligned and inserted in the hole. (c) Object is locked against the hole sides. (d) User pulls the arm away breaking the bound with the object, which falls inside the cube.  }
      \label{fig:ShapeSortingSnapshots}
\end{figure*}
\begin{figure*}[thb]
      \centering
      \includegraphics[width=.9\textwidth, trim= 1.5cm 22cm 1cm 2cm,clip]{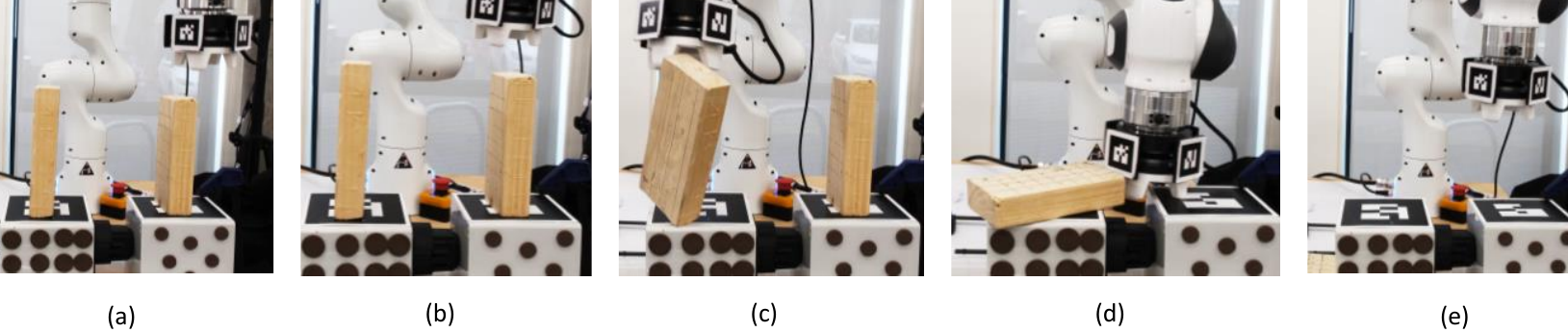}
      \caption{Task's stages for the second experiment: (a) Robot approaches the first obstacle. (b) First obstacle is avoided. (c) Second obstacle is pushed out of the workspace. (d) Second obstacles is pushed out. (e) Robot continues executing the autonomous manipulation task until stopped.}
      \label{fig:ObstAvoid}
\end{figure*}
\autoref{fig:ShapeSortingSnapshots} shows a sequence of frames extracted from the video recorded during the experiments, highlighting the four sub-tasks of this experiment. The experiment was also repeated multiple times to evaluate the robustness and repeatability. Our operator was successful in all four trials showing that the proposed method is robust and has repeatable results. The stability of a communication disconnection between master and replica robot was also successfully tested in one of these tests, validating the robustness and safety of the proposed architectures to an eventual disconnection. It is worth noting that the oscillatory behaviour shown in some of the videos after dropping the cube into the box is not an indication of instability. These motions are caused by releasing the energy accumulated in the non-linear controller spring. They are due to the error between the current pose and the desired pose required to break the bond between object and end-effector. In other words, it is like if we are pulling something with all our strength and suddenly gives up, generating an oscillation of our limbs to release the accumulated energy safely.  

\subsection{Superimposition of Manipulation and Teleoperation}
One of the most challenging tasks for the operator during teleoperation is to move the end-effector across long distances accurately. To solve this issue, we have introduced the possibility of updating the end-effector pose via keyboard inputs in \cite{Babarahmati2021TeleCoop}. This solution allows for repositioning the robot workspace bypassing the haptic interface, and it works well in controlled workspaces. However, it struggles in dynamic scenarios where moving obstacles could interfere with the task. Thus, we thought that we could combine the motion reliability of a pre-planned trajectory with the flexibility of the operator's control by allowing to superimpose to a predefined motion a teleoperation command from the user. To test the feasibility of this application, we set the robot on a circular trajectory on a collision course with two obstacles. Then, the user had to teleoperate the system to avoid colliding with the two objects and try simple manipulation of the obstacles before allowing the robot to hit them. 

\begin{figure*}[thb]
      \centering
      \includegraphics[width=.9\textwidth, trim= 0cm 4.5cm 0cm 4cm,clip]{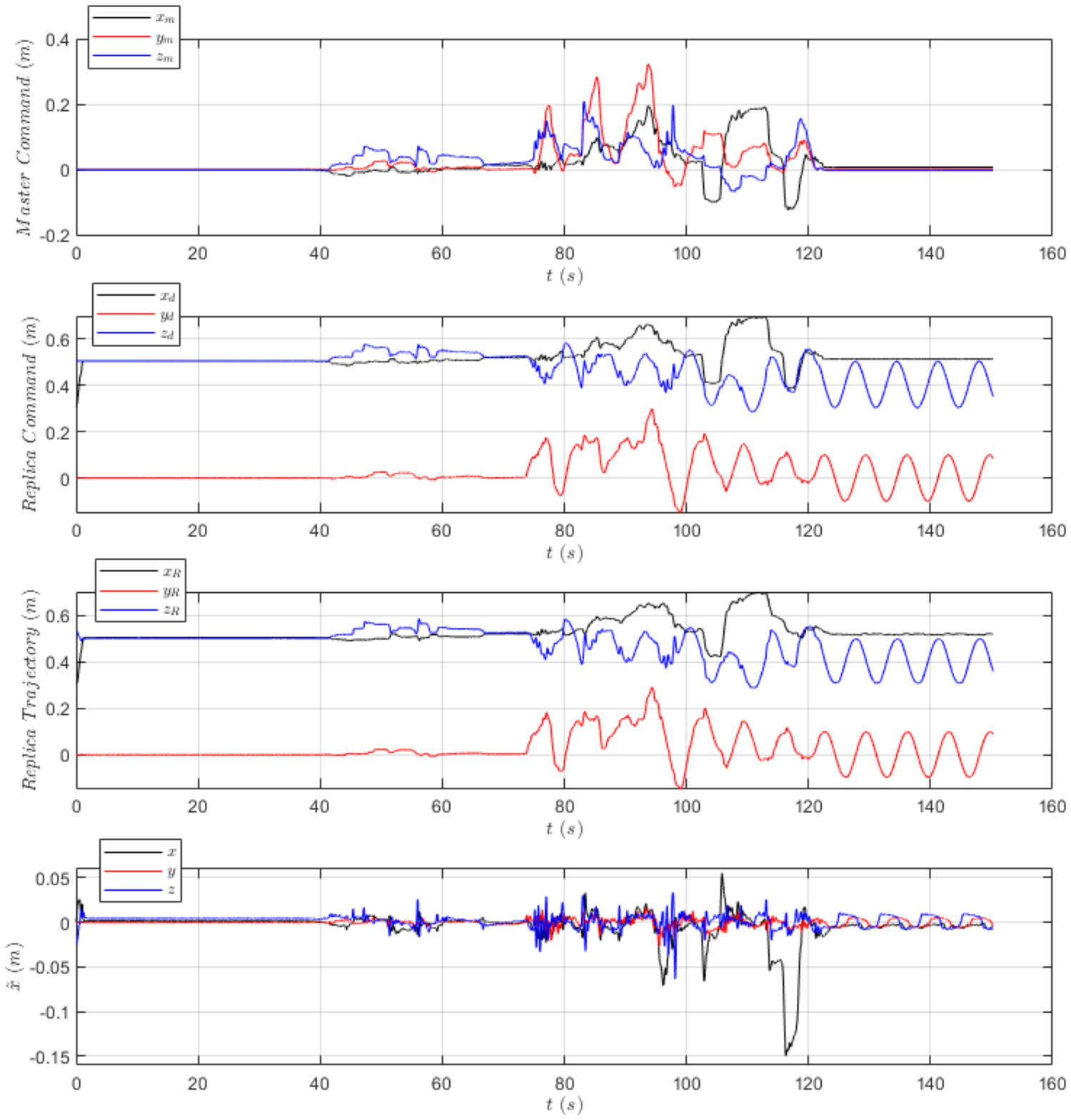}
      \caption{The data show the tracking data from the superimposition experiments.The tracking error shows that the robot is mostly within the desired accuracy $\tilde{x}_b=\SI{0.05}{\meter}$. There are four peaks that exceed the tracking accuracy when the robot is interacting with the environment.}
      \label{fig:SuperimposedTime}
\end{figure*}

\begin{figure}[thb]
      \centering
      \includegraphics[width=\columnwidth, trim= 5cm 8.5cm 5cm 8.8cm,clip]{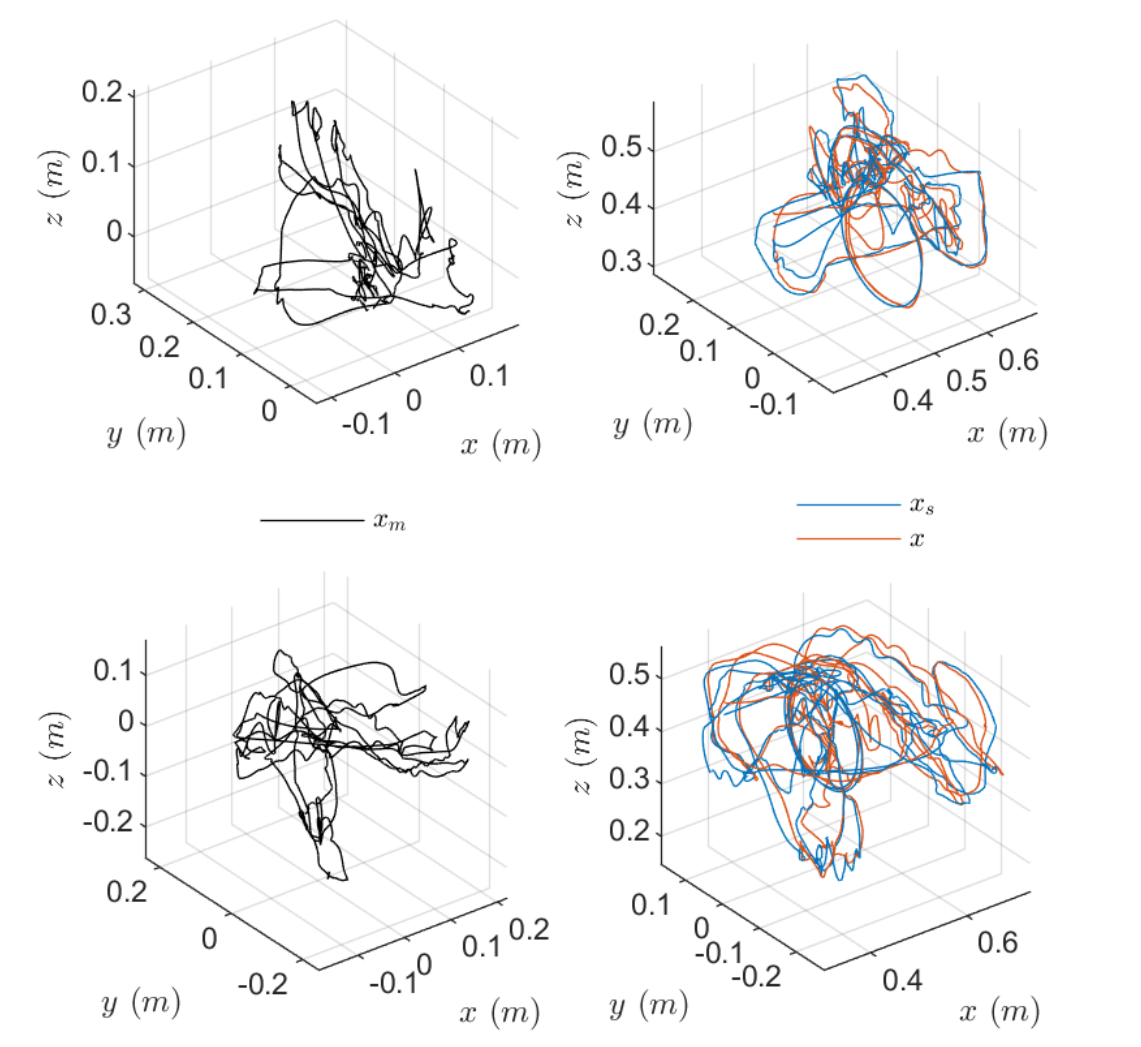}
      \caption{On the top the first trail of the superimposition feasibility experiment. On the bottom the second trial. The effect of the user input ($x_\text{m}$) is evident in both the superimposed trajectories ($x_\text{s}$), and the measured end-efector trajectory. The undisturbed circular trajectory can be still be seen in both trials in the $x_\text{s}$ and $x$ in both cases.}
      \label{fig:Superimposed}
\end{figure}

\begin{figure}[thb]
      \centering
      \includegraphics[width=\columnwidth, trim= 1cm 23.9cm 10cm 1cm,clip]{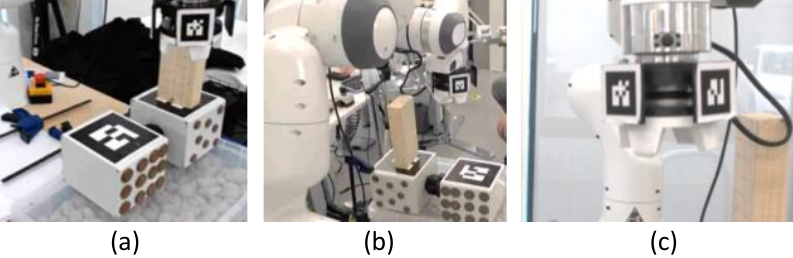}
      \caption{This figure showcases the importance of providing the user with adequate special awareness during dynamics task, which is challenging when dealing with fixed camera placement. (a) and  (b) are the two camera view provided to the user, which do not allow to detect the cable collision. (c) A third camera that allows to identify the collision. }
      \label{fig:3PoV}
\end{figure}

Two trials are conducted for this experiment. The data from the first trial are shown in \autoref{fig:SuperimposedTime}. They indicate that the desired tracking accuracy is respected; however, there are four peaks where it is exceeded due to the interaction with the environment. Nevertheless, the system recovers as soon as the perturbation is removed. A tridimensional visualisation of the trajectory for the master and the replica during these experiment is shown in \autoref{fig:Superimposed}.
We superimpose a periodic circle motion of $\SI{20}{\centi\meter}$ of diameter and with a period of $\SI{5}{\second}$ centred in between the two obstacles. They prove that the superimposition of the teleoperation command to the autonomous trajectory is feasible and does not impact the robustness and safety of interaction of the proposed method, as highlighted in \autoref{fig:ObstAvoid}. However, this experiment also highlights a limited spatial awareness of the operator and its impact on dynamic three-dimensional tasks. The recorded video data indicate that this is partially associated with limited depth perception provided by the cameras; nevertheless, other factors are affecting it. For example, we have found that this is also related to the lack of haptic feedback beyond the end-effector, shown in \autoref{fig:3PoV} where a cable can be observed mounted on the robot touching the obstacle without the operator noticing it.

\section{Discussion \& Conclusion}\label{sec:Discussion}

The experiment on haptic fine teleoperation combined with force manipulation highlights the capabilities of our method to address complex environmental dynamics. This architecture has a decoupled stability between the master and the replica controllers, and it does not rely on the projected dynamics. These two factors determine the method scalability due to its computational efficiency and its robustness to communication disruption, unstructured interaction, and singularities. Additionally, the updated architecture can constraint the motion power within the limits of the replica robot allowing more challenging manoeuvres that are not possible with our earlier implementation of the FIC in teleoperation \cite{babarahmati2020,Babarahmati2021TeleCoop}. 

The first experiment validates that the proposed teleoperation method can be used to perform robust fine manipulation during uncertain dynamic interaction with the environment. The second experiment shows how it is possible to superimpose the teleoperator action and autonomous trajectories. This operational method provides a hybrid method where we can exploit the benefit of autonomous manipulation without renouncing the adaptability to environmental changes provided by the human operator. Manipulation algorithms can provide outstanding repeatability and account for the physical limitation of the robot in both kinematic and dynamic, but they might only partially capture complex environments. Therefore, the human operator can be exploited to respond to the variable components of the task which are difficult to capture using model-based solutions, such as reacting unforeseen obstacles and interacting with unknown objects.

The proposed control architecture showed flexibility of interaction, and a multi-modal deployment capabilities with the possibility to seemingly switch between manipulation and teleoperation as well as superimposing the two modality in presence of communication delays. Nevertheless, our experiments also indicate a limited spatial awareness of the user, which seems to increase with the task' dynamism. The limited perception directly impact the operator's cognitive load \cite{aymerich2017non,makin2017neurocognitive}, which has to learn and adapt his strategies to the system. This should be improved in the future by studying how to improve the embodiment of artificial arms. 

\balance
\bibliographystyle{IEEEtran}
\bibliography{root}
\end{document}